# Predicting Rebar Endpoints using Sin Exponential Regression Model

Jong-Chan Park, Hye-Youn Lim, and Dae-Seong Kang
*Department of Electronic Engineering*
*Dong-A University, Republic of Korea*
pakw2015@naver.com, hylim@dau.ac.kr, dskang@dau.ac.kr

**Abstract**

*Currently, unmanned automation studies are underway to minimize the loss rate of rebar production and the time and accuracy of calibration when producing defective products in the cutting process of processing rebar factories. In this paper, we propose a method to detect and track rebar endpoint images entering the machine vision camera based on YOLO (You Only Look Once)v3, and to predict rebar endpoint in advance with sin exponential regression of acquired coordinates. The proposed method solves the problem of large prediction error rates for frame locations where rebar endpoints are far away in OPPDet (Object Position Prediction Detect) models, which pre-predict rebar endpoints with improved results showing 0.23 to 0.52% less error rates at sin exponential regression prediction points.*

## 1. Introduction

Recently, the expansion of artificial intelligence and robots has accelerated the unmanned manufacturing industry. Artificial intelligence's self-learning capabilities and creative capabilities to create analysis capabilities that are superior to existing statistical analysis are showing innovation in the manufacturing industry.

In rebar processing production, automated smart production systems that minimize loss rates such as automatic calibration technology and load optimization technology are needed. Currently, the calibration time and calibration accuracy of processing rebar factories depend on the proficiency of the workers. Also, rebar processing has problems with quality and safety accidents during processing [1]. Therefore, in order to minimize the defect of processed rebar, research is needed to improve productivity by detecting the end point of rebar. Before the product are released and predicting errors in calibration values. However, the most important factor in the processing of coil rebar is rebar calibration. The technology is not currently automated and standardized.

A system for predicting self-driving motion is being developed using machine learning prediction algorithms as shown in Figure 1.

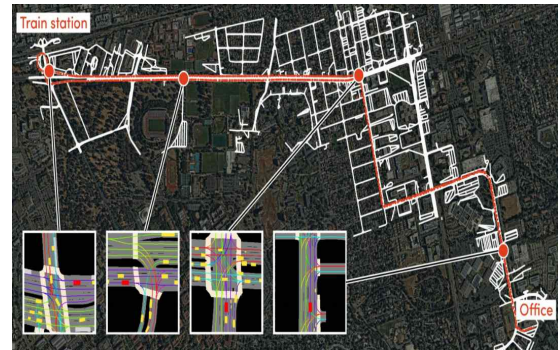

*Figure 1. Visualize self-driving motion prediction*

Previously studied OPPDet [2] is a model for prediction the end points of rebar using non-linear regression. In this model, the prediction ahead of 10 frames was well done, but the prediction error rate of distant frames increased significantly or an unpredictable problem occurred. In this paper, we propose an improved model that predicts endpoints by applying sin exponential regression to improve the problems of these OPPDet models.

## 2. Related Work

### 2.1 YOLO v3

YOLO v3 [3] is an object detection model that guesses the type and location of objects by simply viewing images once. YOLO, a 1-stage detector, has high accuracy and fast detection speed, making it an optimal model because it has to process incoming input images through vision cameras in real time and only needs to consider accuracy and detection speed and detect objects.

YOLO's convolutional neural network architecture transforms a given input image to a fixed size, as shown in Figure 2, and then divides it into an S*S grid. It then passes through a convolutional neural network to output tensors with the shape of S*S* (5*B+C). In this case, S=7 and B=2 are commonly used, with B denoting the number of bounding boxes predicted for each grid, and C denoting the class type. In addition, 5 multiplied by the expression B represents five predicted values, corresponding to the center point of the object (x, y), the length of the object (w, h), and the confidence probability P of the object. Then, for the S*S grid, if the probability values are printed for each class type, the bounding box is calculated based on the probability values [4].

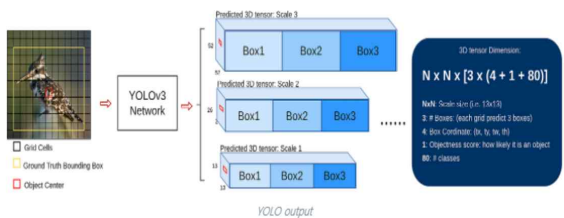

*Figure 2. Yolo neural network structure*

## 2.2 Linear Regression

Linear regression is one of the most basic machine learning techniques, which model data in linear correlations to predict the values we want to know. As shown in Figure 3 below, linear regression is aimed at obtaining a straight-line expression that best represents these data, given N data.

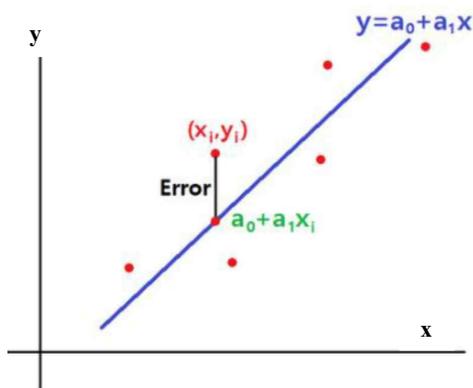

*Figure 3. Linear regression*

## 2.3 Exponential Regression

Exponential regression is to obtain an exponential expression that best represents N data, assuming that N data are given as shown in Figure 4 below. In this paper, we use an algorithm to predict the endpoints of rebar by applying exponential regression functions.

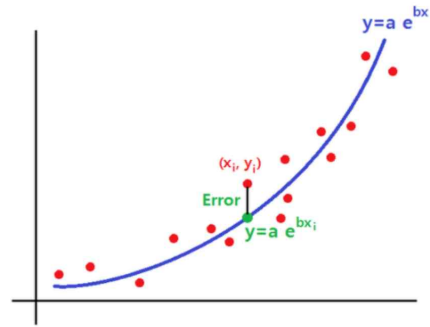

*Figure 4. Exponential regression*

## 3. Proposed Method

First, we build an image dataset divided by frames from rebar endpoint images to proceed with learning datasets on YOLO v3.

Second, the learned YOLO v3 detects the image of the rebar endpoint coming in from the machine vision camera and displays the bounding box in real time. The information obtained from the bounding box (left, top, width, height) produces a time-per-frame (t) axis for the center coordinate (x, y) of the rebar endpoint and the input obtained through the machine vision camera.

Third, the center coordinates (x, y) for the rebar endpoints are two parts: x coordinates (x, t) for the time axis and y coordinates (y, t) for the time axis, and the points (coordinates) are divided into two parts. The points (coordinates) shown in each graph are exponentially regressed to produce a predictive exponential function for the points (expression x for t expression y for t). The sin exponential formula is shown below.

$$y = \exp(ax+b) + \sin(a) \quad (1)$$
$$\log(y) = ax+b + \sin(a) \quad (2)$$

Fourth, to apply sin exponential function expressions, we take x for the time axis and y for the time axis log function. Then, we create a linear regression model and fit the x-coordinate (x, t) for the time axis x and the y-coordinate (y, t) for the time axis.

Fifthly, we obtain coefficients and intercepts via fit, and we obtain x and y values by applying the time value (t) of the desired location for prediction of x-coordinates (x, t) for the time axis and y-coordinates (y, t) for the time axis. The values of x, y are combined into coordinates (x, y) to display them in the grid region and determine whether the coordinates (x, y) are located outside the specific region to determine the defect. The method proposed in this paper is a model that can quickly and accurately analyze the most important prediction points in rebar correction in advance by applying sin exponential regression instead of nonlinear regression in the structure of previously studied OPPDet.

To apply sin exponential regression, we propose an sin exponential regression model that predicts endpoints via the generated predictive sin wave exponential function expression after taking log functions x for the time axis between the fourth and fifth processes and y for the time axis.

## 4. Experiment and Results

The results of this experiment compare the predicted position of sin exponential regression, cos exponential regression, exponential regression, non-linear regression, and the error of the actual position through the proposed method. Through the proposed method, Figure 5 generated a graph of the center coordinates (x, y) as two parts: x-coordinates (x, t) for the time axis and y-coordinates (y, t) for the time axis, and the red line is a function graph for the results of comparing sin exponential regression (A) and cos exponential regression(B) and exponential regression (C) for each graph.

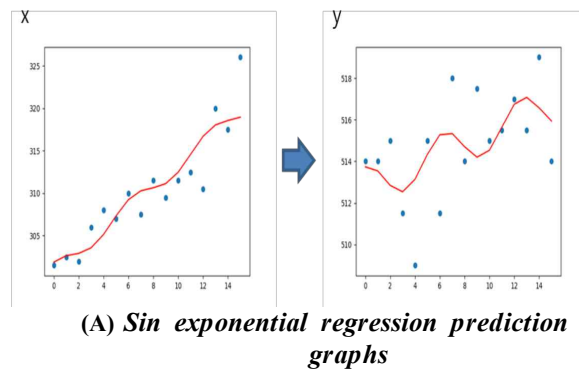

**(A)** *Sin exponential regression prediction graphs*

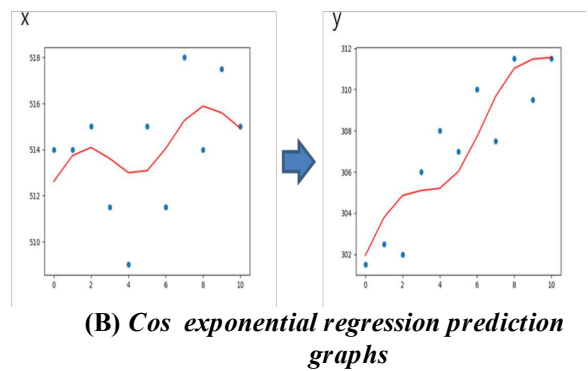

**(B)** *Cos exponential regression prediction graphs*

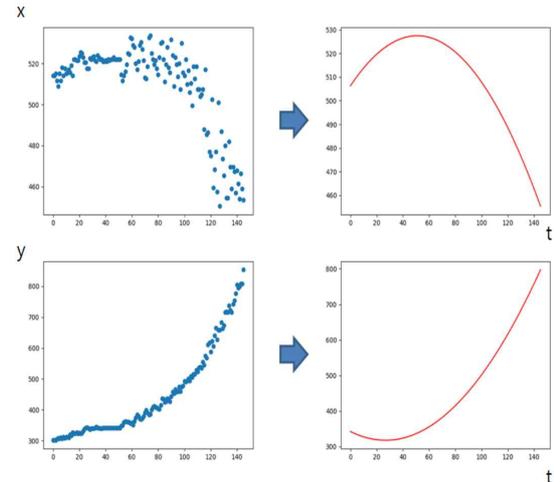

**(C)** *Exponential regression prediction graphs*
*Figure 5. Rebar end-point prediction graphs*

Figure 6 shows the prediction points in front of 60 frames as red dots from the information obtained from the prediction graph, and Figure (A) in Figure 6 visually shows that the prediction positions are almost identical by applying sin exponential regression. Figure (B) and Figure (C) are the results of applying cos exponential regression and exponential regression, and we can see that the predicted position is slightly off than sin exponential regression.

Figure (D) of Figure 6 applies non-linear regression and we can see that the predicted position is significantly out of error.

Table 1 shows the error rate of the actual position of the predicted position of sin exponential regression, cos exponential regression, exponential regression, and non-linear regression.

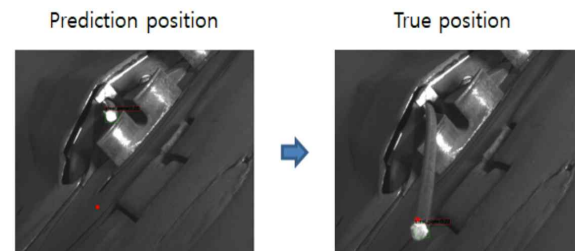

*(A) The predicted location and the actual location of Sin exponential regression*

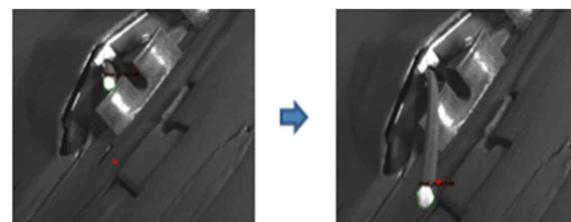

*(B) The predicted location and the actual location of Cos exponential regression*

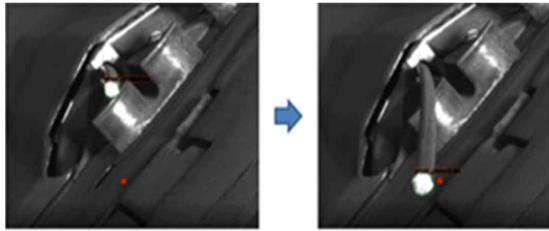

*(C) The predicted location and the actual location of Exponential regression*

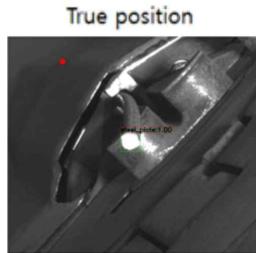

*(D) Actual location of Non-linear regression*
**Figure 6. Comparison of predicted and actual locations of rebar endpoints**

**Table . The error rate between the predicted position and the actual position**

| Regression | Coordinates | Error rate |
|---|---|---|
| Sin exponential regression | x, y | 0.23%. 0.52% |
| Cos exponential regression | x, y | 0.41%, 1.82% |
| Exponential regression | x, y | 0.53%, 2.74% |
| Non-linear regression | x, y | - |

## 5. Conclusion

In this paper, we propose a YOLO v3-based rebar endpoint prediction model with sin exponential regression, cos exponential regression, exponential regression, and non-linear regression. Using the proposed method, we show the most accurate prediction rate by obtaining an error rate of 0.23% at the x-coordinate and 0.52% at the y-coordinate when applying sin exponential regression at the prediction position before 60 frames. In conventional non-linear regression, errors are large or out of frame range. The proposed method improves the problem of poor remote frame position prediction performance in conventional non-linear regression and analyzes prediction points quickly and accurately in advance to present improvements in calibration time and accuracy.

Further research is needed to make predictions while removing unnecessary data due to the severe shaking of rebar.

## Acknowledgment

This work was supported by the National Research Foundation of Korea (NRF) grant funded by the Korea government (NO.2017R1D1A1B04030870).

## References

[1] J. Park and D. Kang, "An Algorithm for the Determination of Twisted Rebar using Feature Matching", *KIIT*, pp. 21-28, Feb 2021.

[2] J. Han and D. Kang, "OPPDet: Object Position Detection Model for Predicting Endpoints of Rebar", KIIT, pp. 135-137, Oct 2020.

[3] J. Redmon and A. Farhadi, "YOLOv3: An Incremental Improvement", CVPR, Apr 2018.

[4] H. Lee, "Fast hand gesture recognition using CNN and edge detection", *Graduate Schhol of Seoul National University*, pp. 1-10, Feb 2018.